\documentclass[letterpaper, 10 pt, conference]{ieeeconf}  % Comment this line out if you need a4paper

\IEEEoverridecommandlockouts                              % This command is only needed if 
\usepackage{graphicx} % for pdf, bitmapped graphics files
\graphicspath{{./figures/}}
\usepackage[font=small]{subcaption}
\usepackage[font=small]{caption}
\usepackage{float}
\usepackage{microtype}
\usepackage{comment}

\usepackage{amsmath} % assumes amsmath package installed
\usepackage{amssymb}  % assumes amsmath package installed
\usepackage{bm} %bold italic vectors
\usepackage{multirow, array}

\usepackage{makecell}
\usepackage{tabularx}
\usepackage{tabulary}
\usepackage{hhline}

\makeatletter
\let\NAT@parse\undefined
\makeatother

\usepackage{algorithm}
\usepackage{algpseudocode}

\usepackage[square, numbers, sort&compress]{natbib}
\usepackage[hidelinks]{hyperref}
\usepackage[capitalise]{cleveref}

\crefname{equation}{}{}
\crefname{figure}{Fig.}{Figs.}

\usepackage{xparse}
\usepackage{xcolor}
\usepackage{diagbox}
\usepackage{booktabs,siunitx,subcaption}
\usepackage{multirow}
\title{\LARGE \bf
Simulator Adaptation for Sim-to-Real Learning of Legged Locomotion via Proprioceptive Distribution Matching
}

\author{Jeremy Dao and Alan Fern% <-this % stops a space
\thanks{*This work is supported by the NSF Grant No. IIS-1849343, DGE-1314109.}% <-this % stops a space
\thanks{All authors are with Collaborative Robotics and Intelligent Systems Institute, Oregon State University, Corvallis, Oregon, 97331, USA. }
\thanks{Email: \{\footnotesize daoje, afern\}@oregonstate.edu.}
}

\begin{document}

\maketitle
\thispagestyle{empty}
\pagestyle{empty}

%%%%%%%%%%%%%%%%%%%%%%%%%%%%%%%%%%%%%%%%%%%%%%%%%%%%%%%%%%%%%%%%%%%%%%%%%%%%%%%%
\begin{abstract}

Simulation trained legged locomotion policies often exhibit performance loss on hardware due to dynamics discrepancies between the simulator and the real world, highlighting the need for approaches that adapt the simulator itself to better match hardware behavior.
Prior work typically quantify these discrepancies through precise, time-aligned matching of joint and base trajectories. This process requires motion capture, privileged sensing, and carefully controlled initial conditions.
We introduce a practical alternative based on proprioceptive distribution matching, which compares hardware and simulation rollouts as distributions of joint observations and actions, eliminating the need for time alignment or external sensing.
Using this metric as a black-box objective, we explore adapting simulator dynamics through parameter identification, action-delta models, and residual actuator models.
Our approach matches the parameter recovery and policy-performance gains of privileged state-matching baselines across extensive sim-to-sim ablations on the Go2 quadruped.
Real-world experiments demonstrate substantial drift reduction using less than five minutes of hardware data, even for a challenging two-legged walking behavior. 
These results demonstrate that proprioceptive distribution matching provides a practical and effective route to simulator adaptation for sim-to-real transfer of learned legged locomotion.

\end{abstract}

%%%%%%%%%%%%%%%%%%%%%%%%%%%%%%%%%%%%%%%%%%%%%%%%%%%%%%%%%%%%%%%%%%%%%%%%%%%%%%%%
\section{INTRODUCTION}

Sim-to-real reinforcement learning (RL) has become the dominant paradigm for training high-performance controllers, particularly for robots with complex morphologies such as legged systems. However, sim-to-real gaps persist and often limit real-world performance, a longstanding challenge in robotics learning research \cite{Zhao2020}. Dynamics randomization (DR) \cite{Peng2018} has been the prevailing solution, with newer works adapting randomization ranges using hardware data \cite{Muratore2021, Ramos2019, Du2021}. Despite these advances, several drawbacks remain.

DR typically randomizes only static model parameters and is therefore limited in the types of dynamics it can represent. It also introduces a trade-off between robustness, performance, and sampling efficiency: wider randomization ranges increase task difficulty and prolong training, while excessively large ranges can lead to overly conservative behaviors \cite{He2024}. Fundamentally, DR does not aim to close the sim-to-real gap, but to make policies robust to it. As tasks grow more complex and gaps widen, DR increasingly shows practical limits, underscoring the need for sim-to-real techniques that incorporate real-world feedback.

The main challenge of sim-to-real solutions is how to best quantify the sim-to-real gap. Traditional system identification executes a known sequence of commands in both simulation and on hardware and directly compares the resulting state trajectories, often using Mean Squared Error (MSE) at each timestep \cite{He2025, Fey2025, Yu2019}. While effective for simpler fixed-base systems, this approach becomes far less practical for complex platforms such as legged robots.

These systems are unstable or sensitive to initial conditions, making precise alignment between simulated and real trajectories extremely difficult. Over longer horizons, small differences quickly cascade into large deviations, and real-world stochasticity can lead to divergent trajectories even under identical actions, rendering direct comparisons ambiguous. Moreover, accurate state-matching requires full ground-truth state measurement, including base pose and velocity, to initialize the simulator consistently, typically via motion capture. This introduces additional noise and imposes infrastructure requirements that many users cannot meet.

% The main challenge of sim-to-real solutions is how to best quantify the sim-to-real gap.
% In traditional system identification, typically some known sequence of commands is executed both in the simulator and on the physical hardware, and the resulting trajectory of states are directly compared. These single step methods are often evaluated using Mean Squared Error (MSE) between each state at each timestep \cite{He2025, Fey2025, Yu2019}. While this approach is feasible for simpler fixed base systems, it becomes far less practical for complex platforms such as legged robots. 
% These systems are often unstable or sensitive to initial states, making precise alignment of simulation and real world trajectories extremely difficult.
% Over longer horizon tasks, small differences quickly cascade to large deviations.
% Real-world stochasticity can produce divergent state trajectories even under identical actions, causing direct comparisons to be ambiguous.
% Moreover, successful state-matching requires full state ground-truth measurements, including base pose and velocity, to initialize the simulator consistently with the real robot. 
% This requires a motion capture system, which introduces additional noise sources and imposes practical requirements that some users cannot meet.

To address these challenges, we introduce a sim-to-real approach that adapts the simulator to better capture the real dynamics without requiring motion capture or state matching metrics. Our method treats sampled proprioceptive observation-action trajectories as distributions and quantifies the discrepancy between simulation and hardware using a distributional metric. This metric serves as an objective for optimizing a parameterized simulator modification function, yielding an updated simulator that more accurately reflects hardware behavior. Finetuning in this adapted simulator subsequently improves real-world transfer and overall performance.

% To address these challenges, we introduce a sim-to-real approach that adapts the simulator to better capture the real dynamics without requiring motion capture or state matching metrics. Our method treats sampled proprioceptive observation-action trajectories as distributions and quantifies the discrepancy between simulation and hardware using a distributional metric. This metric serves as a black-box objective for optimizing a parameterized simulator modification function, yielding an updated simulator more accurately reflects hardware behavior. Finetuning in this adapted simulator subsequently improves real-world transfer and overall performance.

We evaluate this process on several sim-to-sim locomotion tasks using the Go2 quadruped, comparing both matching performance and downstream fine-tuned policy performance against state-matching baselines across multiple simulator modification functions. Even without access to base state information or aligned initial conditions, our distributional metric achieves matching performance comparable to state-matching methods that assume oracle state information.

We further demonstrate its effectiveness across three types of simulator modifications—static parameter identification, an action–delta model, and a residual actuator model, showing that it provides meaningful optimization feedback across diverse dynamics-matching tasks. Among these, the residual actuator model trained with our metric yields the best fine-tuned policy. Finally, in a sim-to-real evaluation, we show that less than five minutes of hardware data is sufficient to optimize a residual actuator model that substantially reduces drift on real hardware after fine-tuning.

\section{RELATED WORKS}

\subsection{Online Adaptation}
Online adaptation methods attempt to update the policy at test time without returning to the simulator to finetune. They typically use incoming real time hardware data to update some estimate of real world properties that are then fed into the control policy. In this way, these methods can be viewed as state estimation problems rather than policy ``adaptation".
For instance, conditioning the control policy on dynamics parameters or other privileged information during training allows for the policy to change its output based on different dynamics \cite{Yu2017, Ji2022}. 
An adaptation module can be trained to estimate these parameters from past robot observations, and during deployment provide inputs to the control policy adapting its behavior.
The adaptation modules can also be trained to infer a latent encoding of privileged dynamics parameters rather than the parameters themselves \cite{Kumar2022, Gu2024, Wang2024, Margolis2022}.
Peng et. al. \cite{BinPeng2020} also utilizes latent encodings but directly searches for the optimal encoding to maximize real-world reward instead of estimating it online. In contrast, Hansen et. al. \cite{Hansen2021} updates and finetunes the dynamics latent module at test time using an inverse dynamics prediction objective.

Although these methods achieve successful sim-to-real transfer, a key limitation of adaptation approaches is that the control policy itself remains unchanged. As a result, its robustness is constrained to the range of dynamics encountered during training. 
% If the adaptation module produces parameters or latent encodings outside this range, performance on real hardware may degrade. 
Since hardware data influences only the policy’s inputs, not the policy itself, these methods ultimately function similarly to standard dynamics randomization, merely selecting the most suitable dynamics model from the training distribution.

\subsection{Offline Finetuning}
Offline finetuning methods aim to improve simulator fidelity using hardware data, modifying the dynamics model rather than the policy. A policy trained in a more accurate simulator is then more likely to transfer successfully to hardware.

The simplest form is system identification (system ID) \cite{Yu2019, Jiang2021, Schwendeman2023, Atkeson1986}, where known actions are executed in both simulation and hardware, and model parameters (e.g., masses, frictions) are tuned until trajectories match. More advanced variants learn state-dependent dynamics models rather than static parameters. For example, neural networks can approximate actuator dynamics that are difficult to model analytically \cite{BinPeng2020, Rudin2022, Bohez2022, deepmind_minitaur, Fey2025, Schwendeman2023, Huang2023, Hwangbo2019}, typically mapping joint states to torques and capturing non-linearities such as friction, gearing, and coupling. However, these approaches often require specialized sensing (e.g., torque or contact measurements) that may not be available on all robots.

Another approach modifies the policy’s action through a learned ``delta action" that corrects simulator behavior \cite{He2025, Karnan2020, Lyu2024, Zeng2020}. These methods generally rely on time-aligned full-state tracking errors and may require motion-capture data \cite{He2025}. In contrast, our method compares just proprioceptive state distributions and thus requires only locally-available joint data without time alignment.

The most general simulator modification is a ``delta dynamics" model, which adds residuals directly to the simulator’s next-state predictions, enabling arbitrary corrections to the dynamics. Such models have been applied to policy learning for simple robots \cite{Saveriano2017} and more commonly to model-based controllers \cite{Heiden2021}. Structure can be imposed by learning external force perturbations instead, which retains simulator structure while allowing broad corrective capability \cite{Gao2024, Sontakke2023}.

SimGAN \cite{Jiang2021} is an example of non-state matching cost function. It trains a GAN discriminator to classify simulation vs. hardware tuples, using the discriminator as a reward to train a state-action dependent parameter model. However, GANs are notoriously unstable and sensitive to hyperparameters, whereas our approach uses a simple, closed-form cost.

Miller et. al. \cite{miller2025} is the closest comparison to our proposed sim-to-real framework and also uses the Wasserstein Distance along with Maximum Mean Discrepencacy (MMD) as the measure of the sim-to-real gap. However, we find that the MMD component is not necessary and that the Wasserstein Distance, which can be expensive to compute for multi-dimensional data, can be approximated efficiently with the average of 1D Wasserstein distances across joint dimensions and features.

\section{SIM-TO-REAL PROBLEM SETUP}
%\section{DISTRIBUTIONAL SIM-TO-REAL}
The goal of our sim-to-real system is to adapt the simulator so that a policy learned in simulation behaves in it the same way it behaves on hardware. We use RL as our policy-learning approach, operating within a parameterized simulator that defines a Markov Decision Process (MDP)
% The goal of our sim-to-real system is to adapt the simulator so that a policy learned in simulation behaves in it the same way it behaves on hardware.  The policy will be learning via RL using a parameterized simulator defining a Markov Decision Problem (MDP)
$\mathcal{M}(\theta) = (S, A, T_{\theta}, d_0, R, \gamma)$, where $S$ and $A$ are the state and action spaces, $T_{\theta}$ is the state-transition function parameterized by $\theta$, and $d_0$ is an initial state distribution. The reward function $R$ and discount factor $\gamma$ specify the optimization objective. Note that the simulator parameters $\theta$ will contain parameters native to the physics simulator and possibly additional parameters that depend on a chosen simulator adaptation approach. Given a policy $\pi$ we let $D(\pi,\theta)=\{\xi^{\text{sim}}_1, \dots, \xi^{\text{sim}}_N\}$ denote a set of $N$ simulated trajectories, where each 
trajectory $\xi^{\text{sim}}_i = \{(s_0, a_0), (s_1, a_1), \dots, (s_H, a_H)\}$ is a sequence of states and policy actions collected over a horizon $H$. Each simulator state decomposes as $s_t = (o_t, p_t)$, where $o_t$ contains the proprioceptive observations available on the real robot, and $p_t$ contains privileged information needed to represent the full world state, such as the 6-DOF base pose in the world frame. 
% trajectory $\xi = \{(o_0, a_0), (o_1, a_1), \dots, (o_H, a_H)\}$ contains sequences of proprioceptive observations and policy actions collected over a horizon $H$. 

Our training process begins using reinforcement learning to optimize a policy $\pi_0$ in $\mathcal{M}(\theta_0)$, where $\theta_0$ are nominal simulator parameters. Next, we execute $\pi_0$ on the physical robot to produce $N$ trajectories $D_{\text{hw}} = \{\xi^{\text{hw}}_1, \dots, \xi^{\text{hw}}_N\}$, where each $\xi^{\text{sim}}_i = \{(o_0, a_0), (o_1, a_1), \dots, (o_H, a_H)\}$ is a sequence of prioceptive observations and policy actions.  In addition, we produce the corresponding simulated trajectories $D_{\text{sim}} = D(\pi_0,\theta_0)$. These paired datasets form the basis for measuring the simulation-hardware discrepancy and guiding subsequent simulator adaptation.  Note that while the simulated trajectories contain full state information, the hardware trajectories contain only proprioceptive observations, which are readily available without motion-capture instrumentation.

% The goal of our sim-to-real system is to modify our original simulation such that our learned policy behaves in it the same way that it does in the real world. Suppose we are given a MDP $(S, A, T(s_t, a_t|\sigma_0), R, \gamma)$ with a state transition function $T$ that is conditioned on some initial set of parameters $\theta_0$. We will then use RL to solve it, e.g. train an initial control policy $\pi_0$ that tries to maximize the expected return. The policy will be then be executed on both the physical hardware and the simulation $T$ using the same $N$ sequences of $H$ command inputs, giving us the two datasets $D_{sim}=(\xi_{sim_0}, \xi_{sim_1}, \dots, \xi_{sim_N})$ and $D_{hw}=(\xi_{hw_0}, \xi_{hw_1}, \dots, \xi_{hw_N})$ where a trajectory $\xi$ contains a sequence of observations and actions $\xi=\{(o_0, a_0), (o_1, a_1), \dots, (o_H, a_H)\}$

To quantify this discrepancy, we define a sim-to-real cost function $C_{\text{S2R}}(D_{\text{sim}}, D_{\text{hw}})$ that measures behavioral differences between the simulated and hardware trajectories. Simulator adaptation consists of optimizing the simulator parameters to reduce this cost:
\begin{equation}
\theta_{1} = \arg\min_{\theta}\; C_{\text{S2R}}(D(\pi_0,\theta),\, D_{\text{hw}}).
\end{equation}
The resulting parameters $\theta_1$ define an updated simulator in which we retrain the policy to obtain a finetuned controller $\pi_1$. This process---collecting trajectories with $\pi_i$, evaluating $C_{\text{S2R}}$, optimizing $\theta_{i+1}$, and retraining $\pi_{i+1}$---can be iterated until the sim-to-real discrepancy is sufficiently small, indicating that the simulator has been adapted to closely match hardware behavior.

Note that this approach can be viewed as form of task-targeted system identification, where the task is specified via the reward function. In particular, the approach does not seek a globally accurate simulator over the entire state and action spaces. Rather, the focus is on learning a simulator adapation that is accurate around the state-action distribution of the learned policy.

In order to instantiate the above sim-to-real adapation process we must specify three key components.
\begin{itemize}
    \item A sim-to-real cost function to empirically measure the sim-to-real gap.
    \item A parameterized simulator adaptation function to update the simulator.
    \item An optimizer of the adaption function to minimize the sim-to-real cost.
 \end{itemize}
The following sections describe our choices for these components. 

% Note that the datasets are generated with the current policy $\pi$ rather than some set of known torques or joint actions. We are not seeking a globally accurate simulator, and rather focus on learning the dynamices of the state space around what the learned policy experiences. 
% Since our end goal is for just our control policy to work on the real hardware, global coverage is not necessary and we do not need to burden ourselves with collecting globally representative data. 
% In effect, this is doing ``policy specific" system identification.

\section{A Distributional Sim-to-Real Approach}

%Adding some initial motivation.
A key difficulty in time-aligned sim-to-real matching for locomotion is that small discrepancies in simulator parameters or initial states cause rollouts to diverge rapidly from hardware trajectories, making pointwise comparisons unreliable. One workaround is to periodically reset the simulator to the real trajectory before divergence becomes too large. For legged locomotion, however, precise temporal alignment and resetting require privileged sensing—such as accurate base pose and tightly synchronized timing—which is often impractical on hardware.

Instead, we treat each trajectory as a collection of samples drawn from the stationary distribution induced by the policy interacting with the system. This stationary distribution acts as an invariant signature of the underlying dynamics: although temporal ordering is removed, long-horizon behavioral characteristics—including gait patterns, contact statistics, actuator lag, and damping effects—remain encoded in its shape. Consequently, comparing simulated and hardware trajectories at the distributional level avoids the need for synchronized initial states and time-aligned rollouts while providing a smoother objective for optimizing simulator parameters.

\subsection{Wasserstein Cost Function}
\label{sec:wass}

While there are multiple ways to measure the difference between empirical distributions, we use the Wasserstein distance due to its non-parametric nature and the fact that it is a proper metric. 
%(i.e., it is symmetric, satisfies the triangle inequality, and is zero only when the distributions match). 
Wasserstein distances are widely used in distribution-matching problems such as generative adversarial learning and have also been applied in sim-to-real transfer \cite{miller2025}. A known drawback, however, is that computing the Wasserstein distance in multiple dimensions requires solving an ``optimal transport'' problem, which can be computationally expensive.

Formally, given two empirical distributions $P$ and $Q$ with $n$ samples $\{x_1,\ldots,x_n\}$ and
$\{y_1,\ldots,y_n\}$ respectively, the Wasserstein distance is defined as
\begin{equation}
    W(P,Q) = \inf_{\eta \in S_n} \sum_{i=1}^n \| x_i - y_{\eta(i)} \|,
\end{equation}
where the infimum is taken over all permutations $\eta$ of the $n$ samples. This formulation becomes
increasingly expensive as the dimensionality of the samples grows.

% While there are multiple ways to measure the difference between distributions, we choose the Wasserstein Distance for its non-parametric nature and that is a proper metric (i.e. is symmetric and obeys the triangle inequality). It is popular in other data matching problems like generative adversarial learning and has been used for sim-to-real purposes before \cite{miller2025}.
% However, the Wasserstein distance can be expensive to compute for multi-dimensional data and involves solving the ``optimal transport problem". Given two empirical distributions $P$ and $Q$ with $n$ samples $x_1, ..., x_n$ and $y_1, ..., y_n$ respectively, the Wasserstein distance is defined as:
% \begin{equation}
% W(P, Q) = \inf_\pi\left(\sum_i\|x_i-y_{\eta(i)}\|\right)
% \end{equation}
% over all permutations $\eta$ of the $n$ samples. 

{\bf 1D Marginal Approximation.}
When the distributions are one-dimensional the Wasserstein distance admits a closed form: it reduces to the $L_1$ distance between the sorted samples, or equivalently, the integral of the difference between their empirical cumulative distribution functions. This can be computed efficiently in $O(n \log n)$ time without solving a transport problem.

Although our data are typically multi-dimensional (e.g., each joint state is a multi-dimensional vector), we approximate the multi-dimensional Wasserstein distance by averaging the 1D Wasserstein distances computed independently along each dimension. For example, when comparing the distribution of joint positions for a robot with $10$ joints, we treat the data as ten separate one-dimensional
distributions, compute the 1D Wasserstein distance for each corresponding joint (e.g., joint $1$ in simulation versus joint $1$ in hardware), and take the average over all joints. 

This 1D marginal approximation is motivated by three considerations. First, it is far more computationally efficient than computing true multi-dimensional Wasserstein distances, enabling frequent evaluations during black-box optimization. Second, it is more sample efficient: reliable estimates of high-dimensional Wasserstein distances require significantly more samples than are practical to obtain from hardware rollouts. Third, and most importantly, the simulator modifications we aim to optimize tend to influence the marginal distributions of joint positions, velocities, and actions far more than the inter-joint correlations. In contrast, inter-joint correlations are largely determined by the robot's morphology, closed-loop controller, and contact structure, and therefore
vary only weakly under the simulator modifications we consider. For this reason, matching marginals provides a computationally efficient and empirically sufficient signal for detecting and correcting sim-to-real dynamics mismatches in practice.

% However, when the dimension of the distributions are 1, the Wasserstein distance has a closed form and just becomes the difference in CDFs of $P$ and $Q$, which can be easily computed. Thus even though our data will typically be multi-dimensional, for example joint positions, we choose to formulate our sim-to-real cost function as the average of 1D Wasserstein distances over the data's dimension. For example, if we are comparing sim and hardware distributions of joint positions and our robot has 10 joints, we will instead treat the distributions as 10 separate distributions each of just single joints. We then compute the 1D Wasserstein distance between corresponding joints (sim joint 1 with hardware joint 1) and take the average over all the joints. 
% % Note that for the rest of this paper when we refer to the "Wasserstein distance" we mean the 1D Wasserstein distance.

% While this throws away any correlation between joints, we find that such information is not necessary for an effective enough measure of the sim-to-real gap. If such inter-joint correlation is truly essential to capturing the dynamics of given task/system, then one can choose to compute the true multi-dimension Wasserstein distance instead, or more closer approximate it with something like the Sinkhorn algorithm (cite?). However, for the tasks shown here this is not needed and thus we consider such experiments out of scope of this work.

Since treating trajectories as unordered samples removes all temporal information, it is important to choose features that implicitly encode meaningful dynamic structure. We therefore compare distributions over joint positions, joint velocities, and policy actions. For each feature type, we compute the averaged 1D Wasserstein distance as described above and sum the three results into the final cost. Joint velocities restore some time-dependent and higher-order information that is lost when using positions alone. While we could, in principle, include accelerations, we find them too noisy in both simulation and hardware to provide reliable matching signals. Policy actions additionally reveal differences in motor- or actuator-level effects, such as velocity-dependent torque limits.

To ensure that each feature contributes comparably to the cost, we normalize all values to lie in $[-1, 1]$. Hardware data define the shift and scale parameters, which we then apply consistently to all simulated samples and keep fixed throughout the sim-to-real adaptation process.

% Since treating sampled trajectories as distributions throws away any time information, it is important to consider what features we will compare using the Wasserstein distance. We choose joint positions, joint velocities, and policy action. The average 1D Wasserstein distance is computed for each feature as described above and then summed to make up the final cost. Including joint velocities is important to include back some notion of time and higher order information and just positions. We could include even higher order information like joint accelerations, but find that such accelerations tend to be too noisy in both simulation and on hardware to have any meaningful sense in matching. 
% Policy actions will be informative in deducing motor level effects, such as position-velocity dependent torque limits.
% % Policy actions are included to capture some sense of ``state, action, next state" dynamics. 

% To ensure that each feature is of the same scale and thus have equal weight on the cost function, we also normalize the data to be between -1 and 1. For each feature dataset, we use the hardware data to define the shift and scaling parameters. The same parameters are then used to normalize all simulation samples and remain fixed throughout the sim-to-real process.

\subsection{Simulator Modifications}

% There are a number of ways the simulator can be modified. We consider 3 different modification types, ranging from simple, structured modifications that are easier to optimize to higher parameter models with more expressitivity. 
% \begin{itemize}
%     \item \textbf{Static Simulator Parameters} Changing joint friction and armature parameters than remain fixed throughout training
%     \item \textbf{Action Delta Model} Following \cite{He2025}, a state dependent neural network outputs residual actions that are added to the original control policy action. The model is feedforward network and takes in the past 3 joint position and velocity observations.
%     \item \textbf{Residual Actuator Model} A state dependent neural network outputs residual torques that are added to the nominal output of the high frequency PD controller. Note that unlike the Action Delta model which runs at the policy rate, the Residual Actuator model runs at the simulation rate. This is similar to \cite{Hwangbo2019, Rudin2022} except the model outputs a torque residual rather than the entire torque command. Like the Action Delta model we use a feedforward network with a 3-long history of position and velocity observations.
% \end{itemize}

First consider the nominal simulator which defines the state-action transition function 
\begin{equation}
\label{eq:nominal_sim}
s_{t+1} = T_{\theta}(s_t, a_t)
\end{equation}
where $\theta$ denotes the model parameters and the control policy produces actions via $a_t=\pi(o_t)$. In our setting, the policy outputs joint position targets, which are converted to torques by a low-level PD controller. As a result, the simulator’s transition dynamics follow the rigid-body equations of motion:
% conditioned on some model parameters $\theta$ and where the action comes from our control policy $a_t=\pi(o_t)$. In our case, the control policy outputs $a_t$ are joint position targets that are passed to a PD controller to output joint torque and thus the transition function follows the rigid body dynamics equation
\begin{align*}
M(q_t)\ddot{q}_t + C(q_t,\dot{q}_t)\dot{q}_t + g(q_t) = \tau_t \\
\tau_t = K_p\big(\pi(o_t) - q_t\big)\;-\;K_d\,\dot{q}_t.
\end{align*}
where $K_p$ and $K_d$ are the proportional and derivative gains of the PD controller, $q_t$ are the joint positions, $\tau_t$ is the torque, $M$ is the mass matrix, $C$ is the Coriolis matrix, and $g$ is the gravitational force vector. In this work, the policy actions are updated at 50Hz and the PD controller operates at 1KHz. 

There are several ways to modify the simulator. We consider three types of modifications, ranging from simple structured parameter adjustments to higher-dimensional models with greater expressivity.
% where $K_p$ and $K_d$ are the $P$ and $D$ gains of the PD controller respectively, $M$ is the mass matrix, $C$ is the Coriolis matrix, and $g$ is the gravitational forces.
% There are a number of ways the simulator can be modified. We consider 3 different modification types, ranging from simple, structured modifications that are easier to optimize to higher parameter models with more expressitivity. 

\textbf{Static Simulator Parameters.}
This modification adjusts fixed physics parameters, such as joint friction and armature, that remain constant throughout training. Concretely, we replace the nominal parameter vector $\theta$ in \Cref{eq:nominal_sim} with an updated fixed vector $\hat{\theta}$, resulting in the modified transition function
\begin{equation}
\label{eq:static_sim_params}
s_{t+1} = T_{\hat{\theta}}(s_t, a_t).
\end{equation}

% \textbf{Static Simulator Parameters} Changing joint friction and armature parameters than remain fixed throughout training. This is simply changing the $\sigma$ that \Cref{eq:nominal_sim} is conditioned on to a different fixed vector $\hat{\sigma}$. This changes our state transition function to
% \begin{equation}
% \label{eq:static_sim_params}
% s_{t+1} = T(s_t, a_t|\hat{\sigma})
% \end{equation}

\textbf{Action–Delta Model.} 
For many systems, simulator inaccuracies manifest not only through incorrect static parameters but also through state-dependent actuation effects (e.g., unmodeled friction, torque limits, or latency). To capture these residual discrepancies, we augment the nominal policy with a learned action correction. A state-dependent neural network $\Delta_{a}(s_t \;|\; \hat{\theta})$ with parameters $\hat{\theta}$ outputs a residual action that is added to the original control action, modifying the transition function to
\begin{equation}
\label{eq:action_delta}
s_{t+1} = T_{\theta}\left(s_t,\, a_t + \Delta_a(s_t \;|\; \hat{\theta})\right).
\end{equation}
Under this modification, the simulator dynamics follow the updated PD control rule
\begin{align*}
\tau_t = K_p\big(\pi(o_t) + \Delta_a(s_t \;|\; \hat{\theta}) - q_t\big) - K_d\,\dot{q}_t,
\end{align*}
where the residual policy $\Delta_a$ adjusts the desired joint targets to bring the simulated behavior closer to hardware. Following \cite{He2025}, we implement $\Delta^a$ as a 2 layer feedforward network with hidden sizes $[8, 4]$ that receives the past three joint-position and velocity observations. In our experiments, each joint is modified with its own separate network that outputs a residual position target in the same units as the original policy action (radians).

% \textbf{Action Delta Model} A state dependent neural network outputs residual actions that are added to the original control policy action. This modifies the transition function to be
% \begin{equation}
% \label{eq:action_delta}
% s_{t+1} = T(s_t, a_t + \pi^{\Delta A}(s_t)|\sigma)
% \end{equation}
% and our dynamics are modified through the updated PD control rule:
% \begin{equation*}
% \tau_t = K_p\big(\pi(o_t) - q_t\big)\;-\;K_d\,\dot{q}_t
% %K_p((\pi(s_t) + \pi^{\Delta A}(s_t)) - q) + K_d(-\ddot{q}) = M(q)\ddot{q} + C(q, \dot{q})\dot{q} + g(q)
% \end{equation*}
% $\pi^{\Delta A}$ acts as a residual policy on top of the original control policy in order to the simulation dynamics closer to real.
% Following \cite{He2025}, we use a feedforward network and takes in the past 3 joint position and velocity observations.

\textbf{Residual Actuator Model.}
Finally, we consider residual errors that cannot be captured by modifying parameters or desired joint targets, such as unmodeled motor dynamics, torque saturation effects, or complex frictional behavior. To account for these discrepancies, we introduce a state-dependent neural network $\Delta_{\tau}\big(s_t \;|\; \hat{\theta}\big)$ with parameters $\hat{\theta}$ that outputs a residual torque added directly to the nominal PD controller output. This results in a dynamics that are conditioned on the actuator residual model $\Delta_{\tau}\big(s_t \;|\; \hat{\theta}\big)$ through an updated PD control rule:
\begin{equation*}
\tau_t = K_p\big(\pi(o_t) - q_t\big) - K_d\,\dot{q}_t + \Delta_{\tau}\big(s_t \;|\; \hat{\theta}\big),
\end{equation*}
where $\Delta_{\tau}$ learns to directly correct torque-level discrepancies between the simulator and hardware at 1KHz. 

% \textbf{Residual Actuator Model} A state dependent neural network outputs residual torques that are added to the nominal output of the high frequency PD controller. This makes the transition function conditioned on a residual actuator model $\pi^{act}(s_t)$
% \begin{equation}
% \label{eq:resid_act_model}
% s_{t+1} = T(s_t, a_t|\sigma, \pi^{act}(s_t) )
% \end{equation}
% which affects the dynamics equation as 
% \begin{equation}
% K_p(\pi(s_t) - q) + K_d(-\ddot{q}) + \pi^{act}(s_t) = M(q)\ddot{q} + C(q, \dot{q})\dot{q} + g(q)
% \end{equation}

Note that unlike the action–delta model, which operates at the policy rate, the residual actuator model is evaluated at the simulation rate and therefore offers greater expressivity. This approach is similar in spirit to \cite{Hwangbo2019, Rudin2022}, with the key difference that our network outputs a torque residual rather than a full torque command. As in the action–delta model, we implement the residual actuator network as a 2 layer feedforward model with hidden size $(8, 4)$ that takes a three-step history of joint-position and velocity. Again, each joint is modified with a separate network which outputs a residual torque in Newton-meters.

% Note that unlike the action delta model which runs at the policy rate, the residual actuator model runs at the simulation rate and thus has more expressitivity. This is similar to \cite{Hwangbo2019, Rudin2022} except the model outputs a torque residual rather than the entire torque command. Like the action delta model we use a feedforward network with a 3-long history of position and velocity observations.

\subsection{Optimization}

With the cost function defined, we obtain a sim-to-real optimization problem that can be solved using a variety of approaches, including search-based, sample-based, or gradient-based methods. However, evaluating the cost requires running the simulator and rolling out full trajectories, which makes gradient-based optimization impractical unless the simulator is differentiable. For this reason, we employ sampling-based black-box optimization, specifically the Covariance Matrix Adaptation Evolution Strategy (CMA-ES) \cite{Hansen16a}, using the \texttt{cmaes} Python implementation from \cite{nomura2024cmaessimplepractical}. CMA-ES has been successfully applied to control and dynamics-identification problems closely related to ours \cite{Dehio2015, Yu2019, sobanbabu2025samplingbased, miller2025}, and it is well suited for the non-smooth, noisy objective landscapes that arise in sim-to-real system identification.

% With the cost function defined, we now have a sim-to-real optimization problem that can be solved using various algorithms, such as search, sample-based, or gradient-based methods.
% Given that computing the cost function requires running the simulation and rolling out a trajectory, direct gradient based methods would require a differentiable simulator.
% Thus we choose to use sampling-based blackbox optimization, specifically Covariance Matrix Adaptation Evolution Strategy (CMAES) \cite{Hansen16a}. We use the \texttt{cmaes} Python library provided in \cite{nomura2024cmaessimplepractical}. CMAES has been successfully applied before to control problems \cite{Dehio2015} as well as system id problems very similar to ours \cite{Yu2019, sobanbabu2025samplingbased, miller2025} and works well in non-smooth optimization landscapes.

\section{RESULTS}
\subsection{Experimental Setup}
\label{sec:exp_setup}
In our experiments, we evaluate the proposed distributional sim-to-real framework across three orthogonal axes: 1) three different source-target scenarios, ranging from simple model-parameter shifts to real hardware experiments; 2) compare alternative sim-to-real discrepancy measures, including our distributional Wasserstein metric and two direct state-matching variants; and 3) our three introduced simulator modification parameterizations of increasing expressiveness. Together,
these combinations enable a systematic comparison of cost functions, modification models, and dynamics gaps. Below we detail each of these axes. 

\subsubsection{Source-Target Scenarios} 

All scenarios use the Unitree Go2 quadruped robot. Training is performed in IsaacLab using PPO as the reinforcement leanring algorithm with the official Go2 USD model provided in the repository. In all three scenarios the controller is trained for velocity-commanded locomotion. In particular, during PPO training, the policy receives random linear and angular velocity commands, and the reward encourages accurate tracking of these commands. 
Note that policies are trained with dynamics randomization of the base mass, joint PD gains, and floor friction coefficient. We refer readers to Appendix I for further training details.
We consider three source-target configurations:

\paragraph*{Model Parameter Shift} This idealized scenario evaluates the ability of the sim-to-real pipeline to recover known physical model parameters. The source model is the nominal Go2 simulator, whose transition function $T_{\theta_0}$ includes no joint friction or armature. Specifically, $\theta_0$ specifies all joint frictions and armatures to be zero, with other simulator parameters remaining at their defaults. The target model $T_{\theta_{\text{targ}}}$ is constructed by introducing hand-chosen friction and armature values for the hip, thigh, and calf joints (six parameters total), producing a controlled and precisely known source-target discrepancy. This scenario therefore provides an idealized setting in which the true parameter shift is known and the resulting dynamics mismatch directly affects the policy's ability to follow commanded velocities. It enables a clean evaluation of how accurately each sim-to-real cost function can recover the ground-truth physical parameters. 

\paragraph*{Spring Joint} This scenario evaluates the ability of the sim-to-real pipeline to handle more complex and non-parametric changes in dynamics that cannot be captured by simple modifications to static model parameters. The source model is again the nominal Go2 simulator with the same $\theta_0$ as above. 
The target model $T_{\theta_{\text{targ}}}$ is created by introducing a virtual spring at the rear-right calf joint, implemented as an additional PD controller with gains $P=75$ and $D=0$. Unlike the Model Parameter Shift scenario, this modification introduces history-dependent torques and joint-coupled dynamics that cannot be represented by a small set of friction or armature parameters. In the corresponding hardware experiment, we replicate this effect by physically attaching a compression spring to the same joint (see \Cref{fig:go2_spring_hw}), though the precise spring stiffness is unknown. This scenario therefore tests whether each sim-to-real method can adapt to a richer, partially unmodeled change in joint dynamics.

\begin{figure}[h]
    \centering
    \includegraphics[width=0.75\columnwidth, trim={9.25cm 0cm 10cm 2cm}, clip]{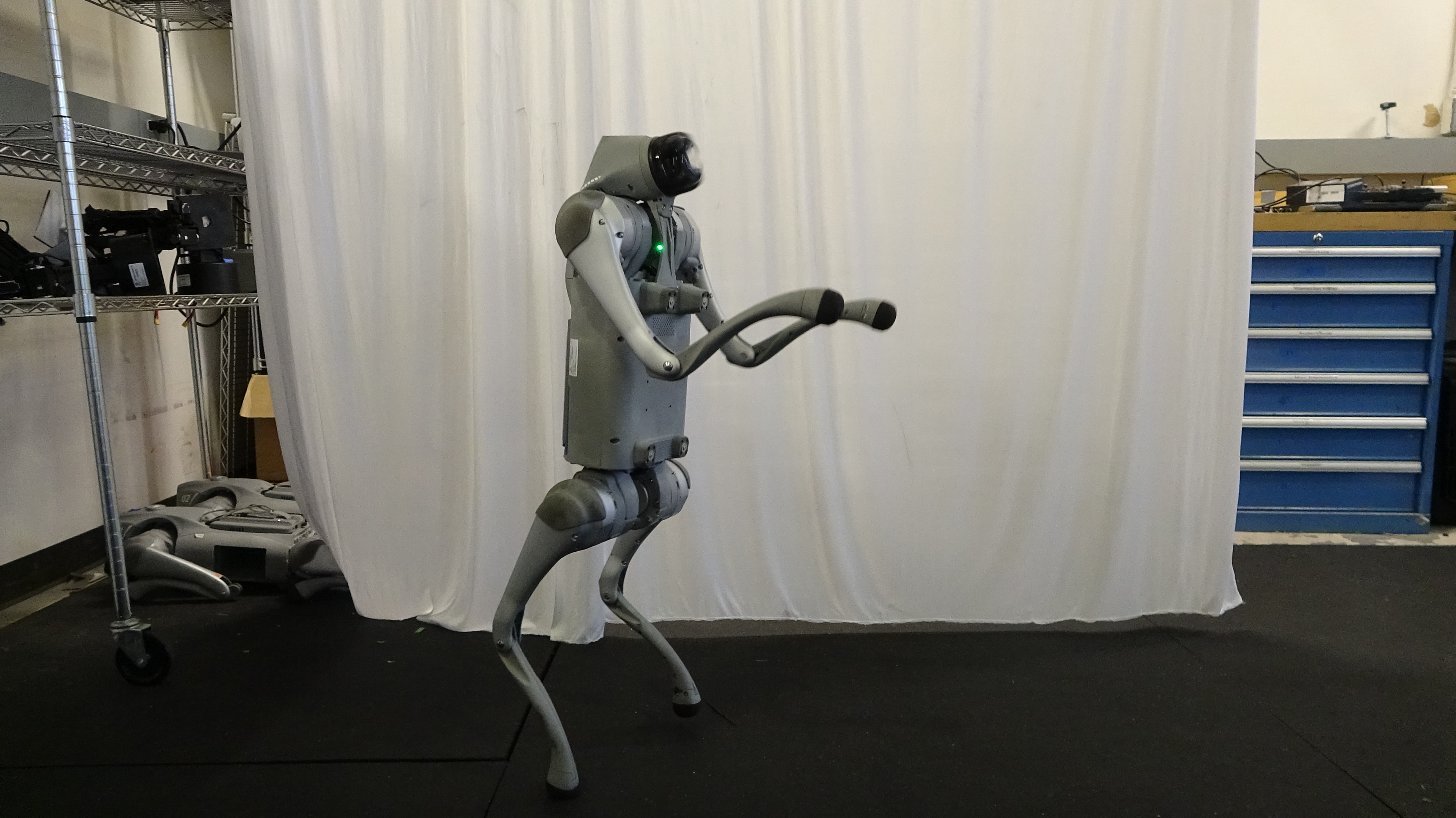}
    \caption{A Unitree Go2 quadruped walking on its hind legs. This is the behavior we attempt to transfer to hardware in our \textit{Bipedal Walking} sim-to-real experiment.}
    \label{fig:go2_2leg_hw}
\end{figure}

\paragraph*{Bipedal Walking} 
This scenario evaluates the sim-to-real pipeline on a challenging task with a substantial dynamics shift and no known ground-truth target parameters. Here the Go2 is required to walk on only its hind legs, which substantially alters contact patterns, load distribution, and overall stability (see \Cref{fig:go2_2leg_hw}). The source model remains the nominal quadrupedal simulator $T_{\theta_0}$. For this scenario we focus on sim-to-real and thus the target is the real-world hardware. 

\subsubsection{Sim-to-Real Cost Functions}

We evaluate three sim-to-real discrepancy measures. Each cost $C_\text{S2R}$ compares trajectories generated by executing the same policy in the source and target models.

\paragraph*{MatchS$(H)$}
This cost function serves as an idealized baseline that assumes access to full state information in the target environment. MatchS$(H)$ measures the direct time-aligned discrepancy between simulated and target trajectories over a finite horizon $H$. During evaluation of this cost, the simulator is reset to the ground-truth target state after every sequence of $H$ steps, enabling matched rollouts over controlled windows of time. The loss is computed as an $L_2$ error over joint positions, joint velocities, base positions, and base velocities, corresponding to the state-matching formulations used in prior work. In our experiments we use $H{=}1$ corresponding to \cite{Fey2025} and $H{=}20$ corresponding to \cite{He2025}). MatchS$(H)$ requires privileged root-state access and strict time alignment—conditions available in simulation
but not on hardware. Increasing $H$ incorporates more temporal structure but also increases the method's sensitivity to trajectory divergence. 

\paragraph*{MatchO$(\infty)$}  
This function evaluates direct $L_2$ discrepancy between proprioceptive observations only (joint positions, joint velocities, and policy actions), without access to privileged root-state information. Because root state is unavailable, the simulator cannot be reset to the target state. Thus the comparison must be performed over an effectively unbounded horizon, making the cost susceptible to the natural divergence that arises between simulated and target trajectories. Unlike
MatchS$(H)$, this loss is hardware-compatible but potentially less stable, as any small modeling error accumulates over time and leads to rapidly increasing $L_2$ penalties.

\paragraph*{Wass} 
This function is our proposed sim-to-real metric based on averaging 1D Wasserstein distances between the simulated and target distributions of proprioceptive observations (see Section~\ref{sec:wass}). This metric does not require time alignment, resetting, or privileged state access. Instead of comparing trajectories pointwise, Wass compares the stationary distributions induced by the policy in each environment, providing a stable and informative signal even in the presence of trajectory divergence.

\subsubsection{Simulator Modifications}

We evaluate three simulator modification parameterizations of increasing expressiveness. Each parameterization defines how the simulator dynamics are altered during the optimization of $\theta$ in the sim-to-real loop.

\begin{itemize}
    \item \textit{FricArm}: Modifying static joint friction and armature parameters
    \item \textit{ActionDelta}: An action delta model, which applies a residual action on top of the policy output. 
    \item \textit{ResidAct}: A residual actuator model, which applies a residual torque on top of the PD controller output. Note that this residual actuator model runs at the same rate as the PD controller.
\end{itemize}

\subsubsection{Experiment Protocol}

For all sim-to-sim experiments, we first train a velocity-commanded locomotion policy $\pi_0$ using PPO in the nominal simulator. We then execute $\pi_0$ in the target simulator to obtain $D(\pi_0,\theta_{\text{targ}})$, consisting of 64 trajectories of 4 seconds each with random linear and angular velocity commands. These trajectories serve as the reference data for sim-to-sim adaptation.

We then estimate the simulator modification parameters by optimizing the selected parameterization $\theta$ with respect to the chosen cost function $C_{\text{S2R}}$:
\begin{equation}
\label{eq:sim2real_opt}
\hat{\theta} = \arg\min_{\theta} \; C_{\text{S2R}}\!\left(D(\pi_0,\theta),\, D(\pi_0,\theta_{\text{targ}})\right)
\end{equation}
Each evaluation of $C_{\text{S2R}}$ during optimization is computed from a freshly generated set of 64 four-second trajectories $D(\pi_0,\theta)$ obtained by running $\pi_0$ in the modified simulator.
We use CMAES to do the optimization, and refer readers to Appendix II for hyperparameter details.
After identifying $\hat{\theta}$, we perform a fine-tuning phase in which PPO is used to reoptimize the policy in the modified simulator $\mathcal{M}(\hat{\theta})$, yielding a finetuned controller $\pi_1$. This process is applied independently for each combination of source--target scenario, simulator modification parameterization, and sim-to-real cost function.

\begin{table}[h]
\centering
\setlength{\tabcolsep}{3pt}
\begin{tabular}{|c|c|c|c|c|c|c|}
\hline
 & \makecell{Hip\\Friction} & \makecell{Thigh\\Friction} & \makecell{Calf\\Friction} & \makecell{Hip\\Armature} & \makecell{Thigh\\Armature} & \makecell{Calf\\Armature}\\
\hline
\makecell{True\\Values} & 0.2 & 0.15 & 0.1 & 0.1 & 0.05 & 0.075\\
\hline
\makecell{Wass} & 0.21 & 0.153 & 0.116 & 0.1001 & 0.0496 & 0.0741\\
\hline
\makecell{MatchS(1)} & 0.2 & 0.15 & 0.09999 & 0.1 & 0.04999 & 0.075\\
\hline
\makecell{MatchS(20)} & 0.1999 & 0.15001 & 0.0999 & 0.0999 & 0.04999 & 0.07499\\
\hline
\makecell{MatchO($\infty$)} & 0.1999 & 0.15001 & 0.09999 & 0.09998 & 0.04997 & 0.07501\\
\hline
\end{tabular}
\caption{Optimized Joint Friction and Armature Values with different cost functions under idealized target environment conditions.}
\label{Table:fricarm}
\end{table}

\subsection{Sim-to-Sim Experiments}

In this section we first consider the accuracy of identifying known model parameters and then evaluate sim-to-sim fine-tuning performance for known and unknown parameter scenarios. 

\textbf{Parameter Identification: Ideal Conditions.} Our first sim-to-sim evaluation is the \textit{Model Parameter Shift} scenario under ideal conditions where there is no noise in the target simulator. The goal is to recover known joint friction and armature parameters across groups of joints. This setting isolates the parameter-identification component of our framework and evaluates its matching efficacy. We optimize joint and friction values using each of the cost functions (MatchS, MatchO, and Wass) and report the results in  \Cref{Table:fricarm}. 
All cost functions successfully find the correct model parameters within numerical rounding error. As expected the time-aligned maching approaches (MatchS and MatchO) perform well under this idealized setting, providing almost perfect parameter identifications. However, we see that the distributional Wasserstein cost function is nearly as accurate with practically negligible errors. These results illustrate that our proposed 1D Wasserstein cost is as effective as state-matching metrics for quantifying the sim-to-sim gap, while providing meaningful optimization feedback without requiring privileged base information or time-aligned trajectories.

\begin{figure*}[t!]
    \centering
    %--- Subfigure 1 ---
    \begin{subfigure}{0.32\textwidth}
        \centering
        \includegraphics[width=\linewidth]{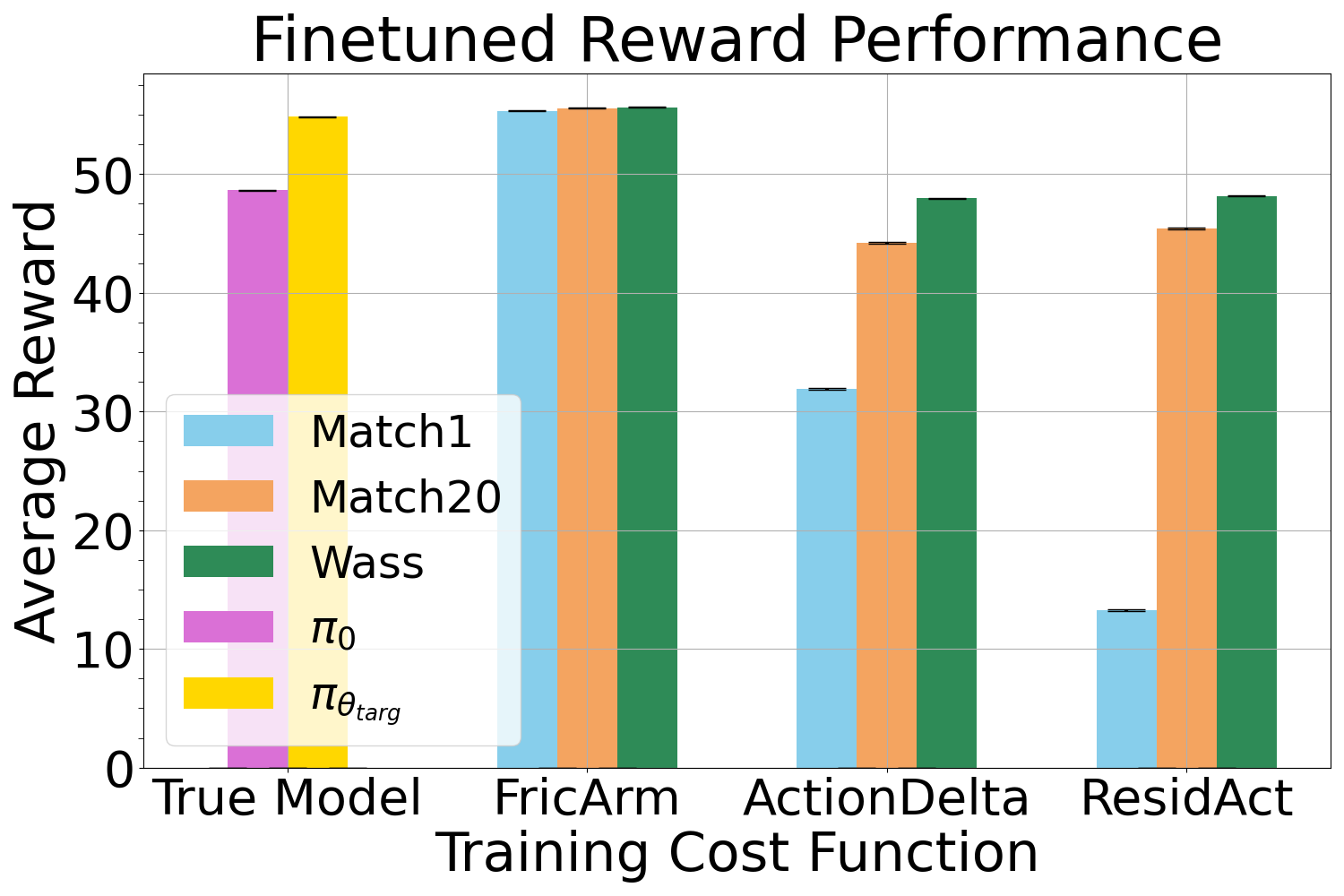}
        \caption{Average sum of rewards over a four second trajectory. Higher is better.}
        \label{fig:fricarm_s2s_rew}
    \end{subfigure}
    \hfill
    %--- Subfigure 2 ---
    \begin{subfigure}{0.32\textwidth}
        \centering
        \includegraphics[width=\linewidth]{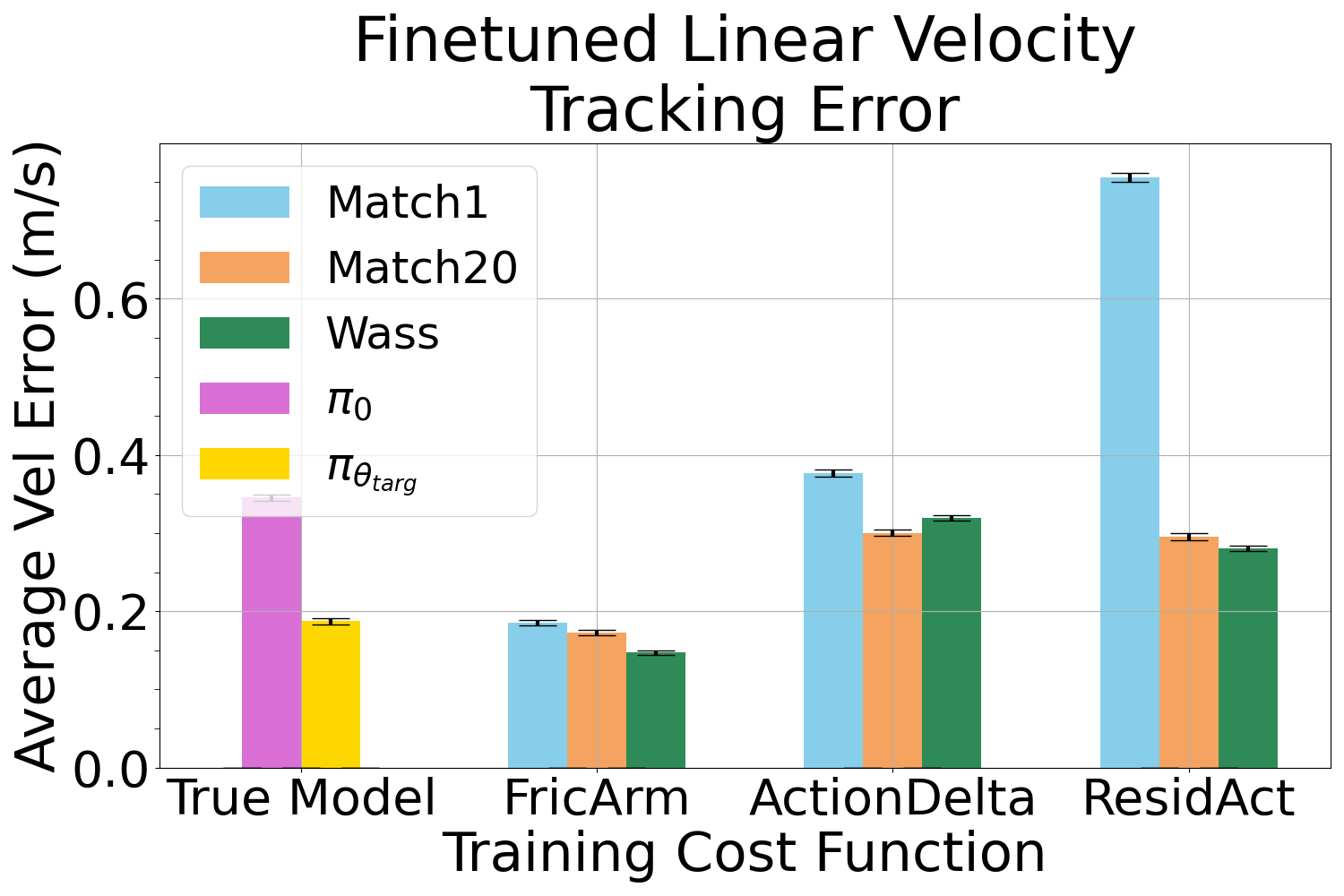}
        \caption{Average linear velocity tracking error. Lower is better.}
        \label{fig:fricam_s2s_linvel}
    \end{subfigure}
    \hfill
    %--- Subfigure 3 ---
    \begin{subfigure}{0.32\textwidth}
        \centering
        \includegraphics[width=\linewidth]{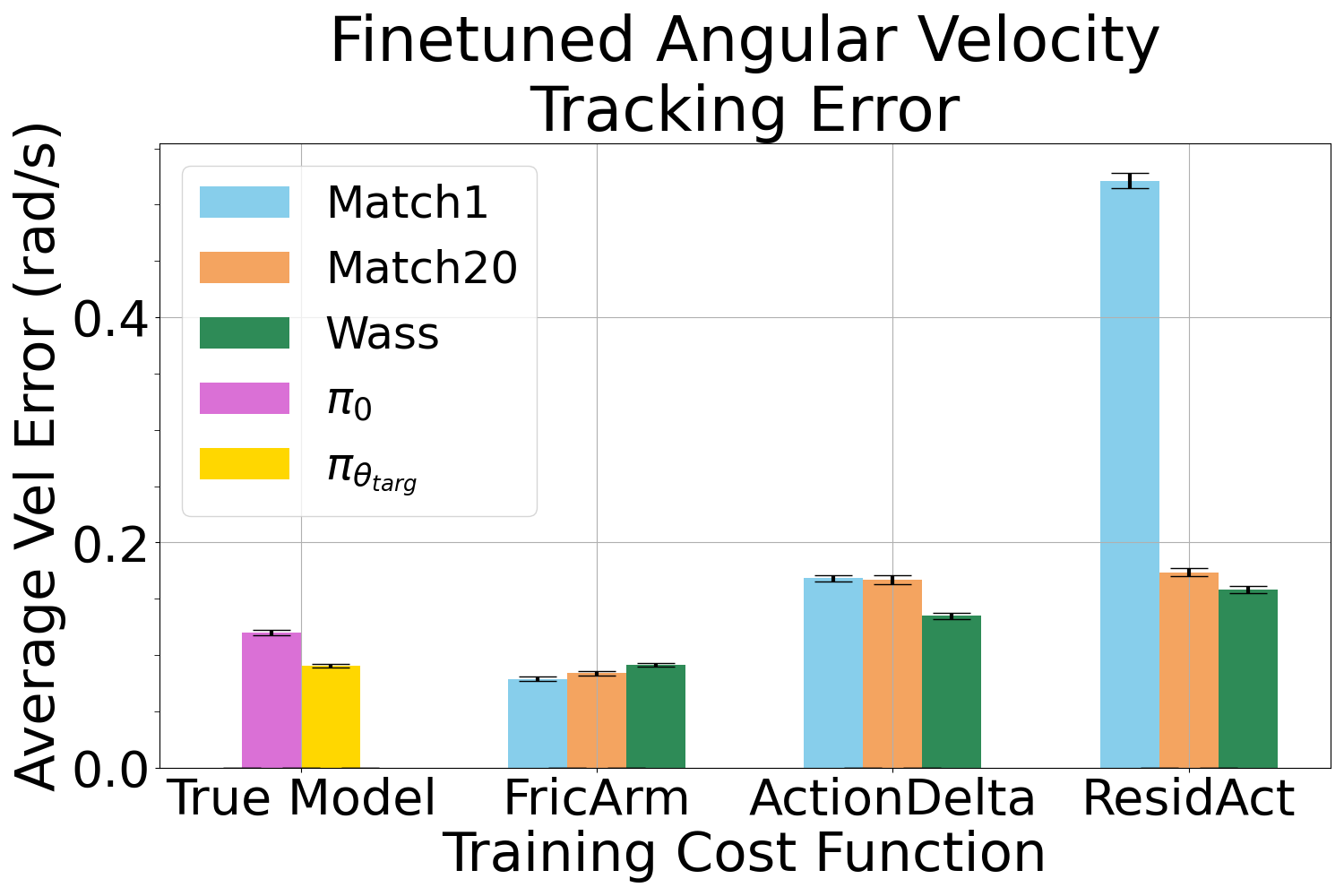}
        \caption{Average angular velocity tracking error. Lower is better.}
        \label{fig:fricarm_s2s_angvel}
    \end{subfigure}

    \caption{Finetuned policy performance for the \textbf{Model Parameter Shift} sim-to-sim test scenario. We average values over 10,000 trajectories with random commands. Error bars represent a 95\% confidence interval.}
    \label{fig:fricarm_s2s}
\end{figure*}

\textbf{Parameter Identification: Noisy Conditions.} We next evaluate the robustness of the sim-to-real costs under conditions that more closely reflect
real-world hardware execution. In sim-to-real settings, target data often contains unmodeled actuator effects, imperfect state alignment, communication delays, and the absence of privileged base-state information. To emulate these effects, we generate noisy target data $D(\pi_0, \theta_{\text{targ}}, \sigma)$ by injecting zero-mean Gaussian torque noise $\mathcal{N}(0,\sigma)$ into the actuator commands and by introducing a random temporal shift of
0.02--0.1 seconds into the logged trajectories. These perturbations simulate common sources of hardware mismatch such as latency-induced torque inconsistencies, actuator heating, and desynchronization caused by sensor timing inaccuracies. Under these conditions, we compare three costs: MatchS$(\infty)$, MatchO$(\infty)$, and Wass. MatchS$(\infty)$ is included as a privileged reference, assuming full access to base position and velocity and representing the best-case performance of state matching when mocap-quality information is available.

\Cref{Table:fricarm_noise} reports the percentage errors for each actuator parameter and the overall average under increasing noise levels. We first compare Wass to MatchO$(\infty)$, both of which do not rely on privileged state information. Across all noise levels, Wass significantly outperforms
MatchO$(\infty)$, with the gap widening as noise increases. In particular, at the highest noise level, mean parameter errors are $18.7\%$ for Wass compared to $86.3\%$ for MatchO$(\infty)$. 

Next, we observe that MatchS$(\infty)$ also provides a substantial improvement over MatchO$(\infty)$ at all noise levels, highlighting the value of privileged base-state information when motion capture is available. However, while MatchS$(\infty)$ performs slightly better than Wass at low noise levels, Wass is competitive at the smallest noise level and significantly better at the two larger noise levels. This demonstrates that the distributional Wasserstein approach remains more robust even in regimes where privileged state access would otherwise be beneficial, making it a strong practical alternative to motion-capture instrumentation for parameter identification in legged systems.

{
\begin{table*}[t!]
\centering
\renewcommand{\arraystretch}{1.4}
\begin{tabular}{|c|c|c|c|c|c|c|c|c|}
\hline
Noise Level & Cost Function & \makecell{Hip\\Fric} & \makecell{Thigh\\Fric} & \makecell{Calf\\Fric} & \makecell{Hip\\Arm} & \makecell{Thigh\\Arm} & \makecell{Calf\\Arm} & Average Error\\
\hline
\multirow{3}{*}{$\sigma=1.0$} 
    &$MatchS(\infty)$ & 0.450\% & 3.787\% & 24.820\% & 7.400\% & 2.600\% & 4.504\% & 7.2602\\
    \cline{2-9}
    &$MatchO(\infty)$ & 17.130\% & 6.047\% & 21.220\% & 5.317\% & 33.800\% & 7.665\% & 15.1965\%\\
    \cline{2-9}
    &Wass & 0.975\% & 11.673\% & 6.680\% & 6.000\% & 15.028\% & 7.467\% & 7.9705\% \\
\hline
\hline
\multirow{3}{*}{$\sigma=2.5$} 
    &$MatchS(\infty)$ & 6.160\% & 10.147\% & 73.940\% & 0.830\% & 1.924\% & 0.440\% & 15.5735\%\\
    \cline{2-9}
    &$MatchO(\infty)$ & 26.635\% & 19.307\% & 101.700\% & 19.846\% & 61.294\% & 15.561\% & 40.7238\%\\
    \cline{2-9}
    &Wass & 16.380\% & 1.540\% & 20.441\% & 5.972\% & 1.538\% & 4.172\% & 8.3405\\
\hline
\hline
\multirow{3}{*}{$\sigma=5.0$} 
    & $MatchS(\infty)$ & 9.990\% & 16.220\% & 178.940\% & 1.630\% & 16.336\% & 3.765\% & 37.8135\%\\
    \cline{2-9}
    & $MatchO(\infty)$ & 17.550\% & 13.333\% & 363.310\% & 33.660\% & 51.420\% & 38.389\% & 86.2770\%\\
    \cline{2-9}
    & Wass & 30.065\% & 1.233\% & 46.567\% & 14.191\% & 13.098\% & 7.345\% & 18.7498\%\\
\hline
\end{tabular}
\caption{Percent error in optimized Joint Friction and Armature values with different cost functions on noisy data. $\sigma$ represents the standard deviation (in $N\cdot m$ of Gaussian torque noise applied at each simulatino step.}
\label{Table:fricarm_noise}
\end{table*}
}

\begin{figure*}[t!]
    \centering
    %--- Subfigure 1 ---
    \begin{subfigure}{0.32\textwidth}
        \centering
        \includegraphics[width=\linewidth]{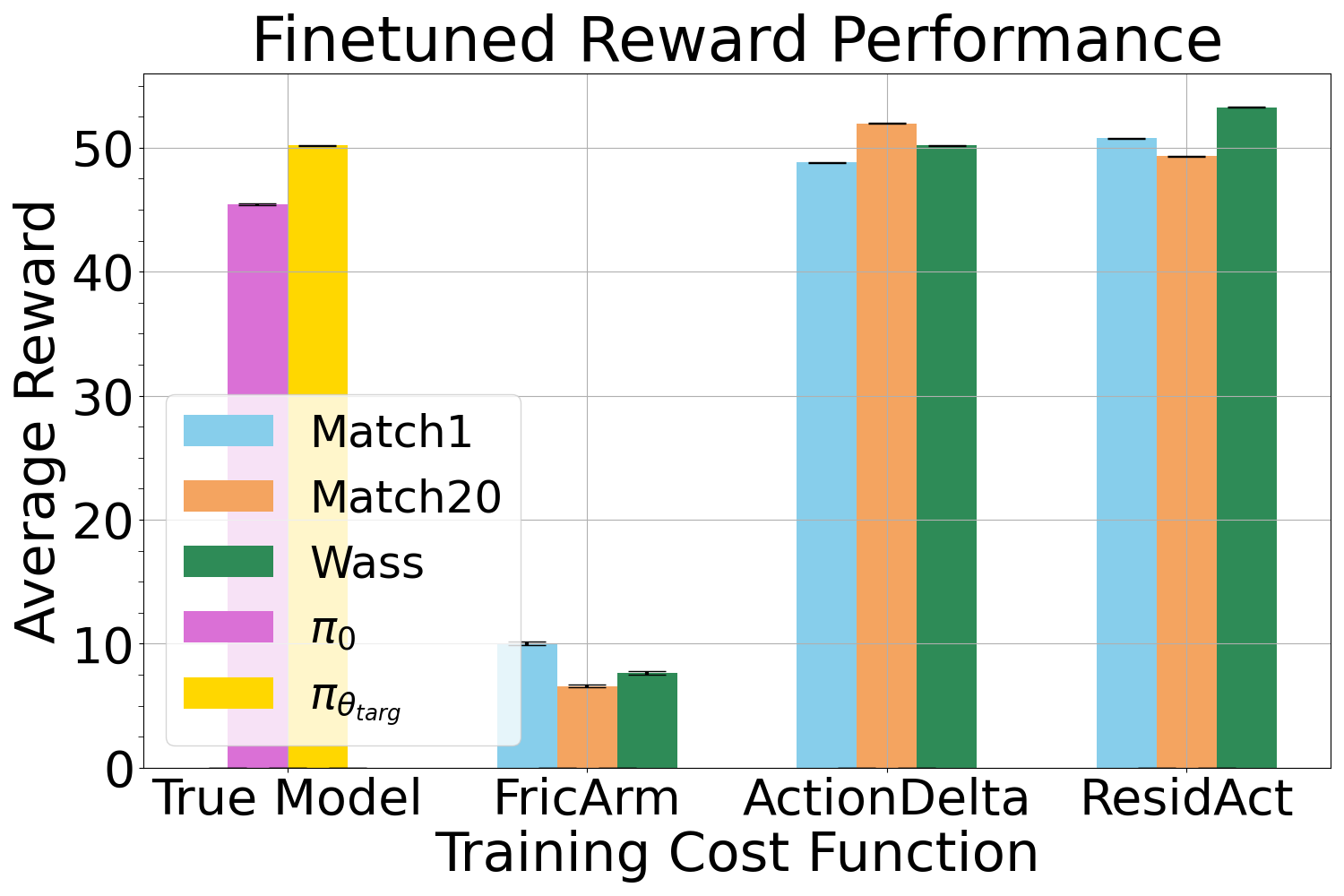}
        \caption{Average sum of rewards over a four second trajectory. Higher is better.}
        \label{fig:spring_s2s_rew}
    \end{subfigure}
    \hfill
    %--- Subfigure 2 ---
    \begin{subfigure}{0.32\textwidth}
        \centering
        \includegraphics[width=\linewidth]{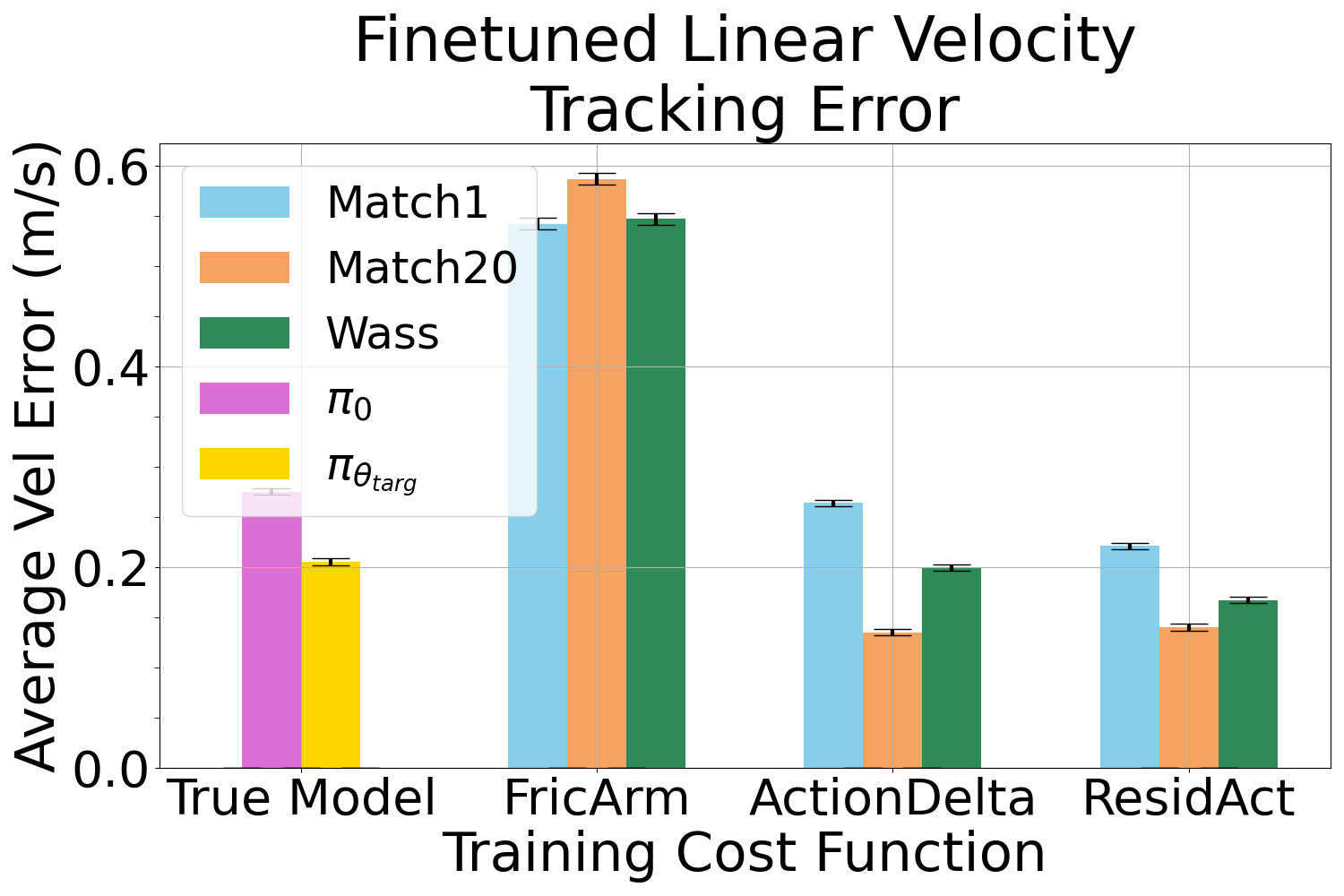}
        \caption{Average linear velocity tracking error. Lower is better.}
        \label{fig:spring_s2s_linvel}
    \end{subfigure}
    \hfill
    %--- Subfigure 3 ---
    \begin{subfigure}{0.32\textwidth}
        \centering
        \includegraphics[width=\linewidth]{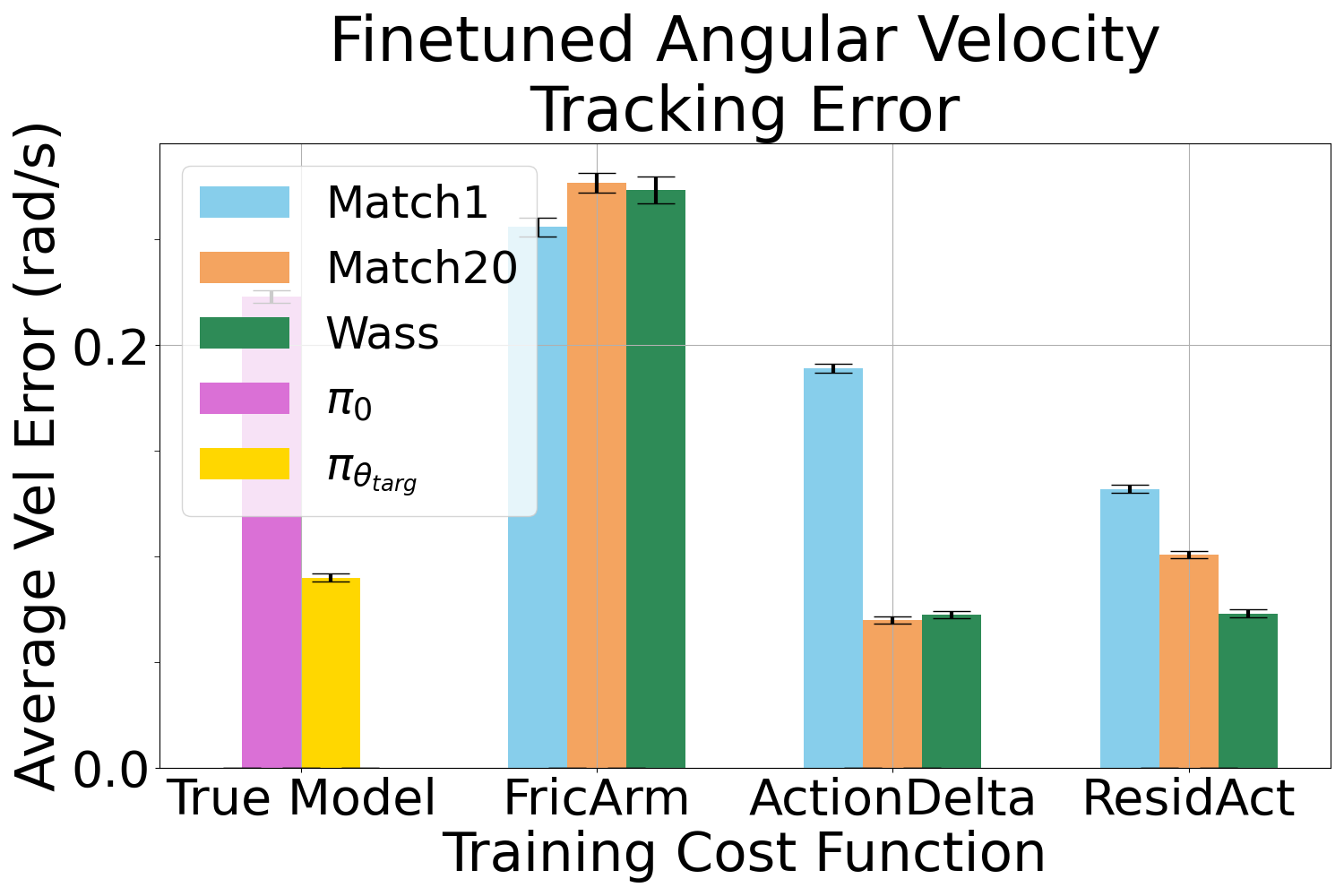}
        \caption{Average angular velocity tracking error. Lower is better.}
        \label{fig:spring_s2s_angvel}
    \end{subfigure}

    \caption{Finetuned policy performance for the \textbf{Spring Joint} sim-to-sim test scenario. We average values over 10,000 trajectories with random commands. Error bars represent a 95\% confidence interval.}
    \label{fig:spring_s2s}
\end{figure*}

\textbf{Sim-to-Sim Finetuning: Model Parameter Shift}. We now consider finetuning in the Model Parameter Shift scenario with no noise. The primary goal is to compare the different simulator modification functions in this idealized scenario. We first use the friction and armature parameters optimized by each of the three cost functions under non-noise scenerios to finetune policies. Each optimized parameter set defines a new MDP, yielding $M(\theta_\text{MatchS(1)})$, $M(\theta_\text{MatchS(20)})$, and $M(\theta_\text{Wass})$. We finetune the initial policy $\pi_0$ in each modified simulator to obtain $\pi_\text{MatchS(1)}$, $\pi_\text{MatchS(20)}$, and $\pi_\text{Wass}$. Each resulting policy is then evaluated on the target model by sampling 10,000 four-second long trajectories with random velocity commands and computing reward and velocity-tracking error. As baselines, we evaluate both $\pi_0$ and $\pi_{\theta_\text{targ}}$, a policy finetuned directly on the target model. 
These two baselines provide, respectively, a lower bound on performance degradation due to the sim-to-sim gap and an upper bound on the performance recoverable by finetuning.
% This will give us a baseline of how much performance is lost because of the sim-to-sim gap and an upper ceiling of much performance can be gained back. 

The Model Parameter Shift finetuning results in \Cref{fig:fricarm_s2s} show that all cost functions result in comparable finetuned performance, closely matching the performance achieved by finetuning directly on the true model. This is expected, as all costs recover the ground truth parameter within neglible error. 

We further conduct the same sim-to-sim procedure for the Model Parameter Shift scenario using the ActionDelta and ResidAct simulator modification functions. The results in \Cref{fig:fricarm_s2s} indicate that these more expressive models lead to substantially worse finetuning performance. Even in the best case, the Wass cost function yields performance no better than the original policy, effectively providing no benefit. This occurs because the ActionDelta and ResidAct networks represent high-dimensional modification spaces that are substantially harder to optimize than the six-dimensional FricArm model. Their greater expressiveness also makes them more sensitive to parameter perturbations, complicating exploration and increasing the likelihood of the optimization becoming trapped in local minima that correspond to small parameter magnitudes and minimal simulator modification.

Notably, the MatchS(1) cost performs particularly poorly. A horizon of one step captures too little dynamical information for high-dimensional neural modification models, despite being adequate for the low-dimensional Model Parameter Shift task. During training, the simulator is reset at every step when using MatchS(1), which encourages the learned models to rely on short horizon cues and makes them prone to producing out-of-distribution outputs when rolled out for longer horizons like during finetuning. This actually causes a larger sim-to-sim gap than was originally present, and consequently leads to finetuned policies that perform worse than the original $\pi_0$.

\textbf{Sim-to-Sim Finetuning: Spring Joint}.
Our second sim-to-sim finetuning experiment is conducted on the \textit{Spring Joint} scenario. As before, we optimize each of the three simulator modification functions using each of the three cost functions, then fine-tune $\pi_0$ in the resulting modified simulators, and evaluate the learned policies on the true spring model. Note that since this scenario only modifies the right rear calf joint of the robot, unlike in the other two scenarios, each modification function only modifies that single joint.

The results in \Cref{fig:spring_s2s} underscore the need for more expressive modification functions than just static model parameter shifts. 
Because the spring mechanism induces complex nonlinear effects that cannot be captured through friction and armature adjustments alone, all Model Parameter Shift variants fail catastrophically, producing dramatically lower reward and velocity-tracking accuracy than the original policy.
In fact, the CMA-ES optimization becomes unstable, continually increasing its exploration variance as it fails to discover any meaningful relationship between the parameters and the matching cost.
This highlights the inadequacy of static parameter shifts for such dynamics and motivates the use of higher-capacity modification models.

By contrast, the ActionDelta and ResidAct models are able to learn useful modifications and yield consistent performance improvements over the baseline policy. As in the previous scenario, the MatchS(1) cost underperforms across all models due to its limited horizon. The Wasserstein-based models again match the performance of the MatchS(20) variants while requiring neither base-state information nor time-aligned trajectories for state resets. After only a single fine-tuning iteration, the Wasserstein metric effectively closes the sim-to-sim gap, achieving performance on par with finetuning directly on the ground-truth simulator.

Across all of our sim-to-sim experiments, we demonstrate that the proposed \textit{Wass} cost function achieves parameter identification and finetuning performance comparable to privileged state-matching baselines, despite not requiring time-aligned trajectories or privileged base information. Moreover, we show that it effectively supports multiple classes of simulator-modification models, highlighting its general applicability as a robust and informative measure of the sim-to-real gap.

\subsection{Sim-to-Real Experiments}

\begin{figure}[h]
    \centering
    \includegraphics[width=0.75\columnwidth, trim={8.25cm 2cm 8cm 4cm}, clip]{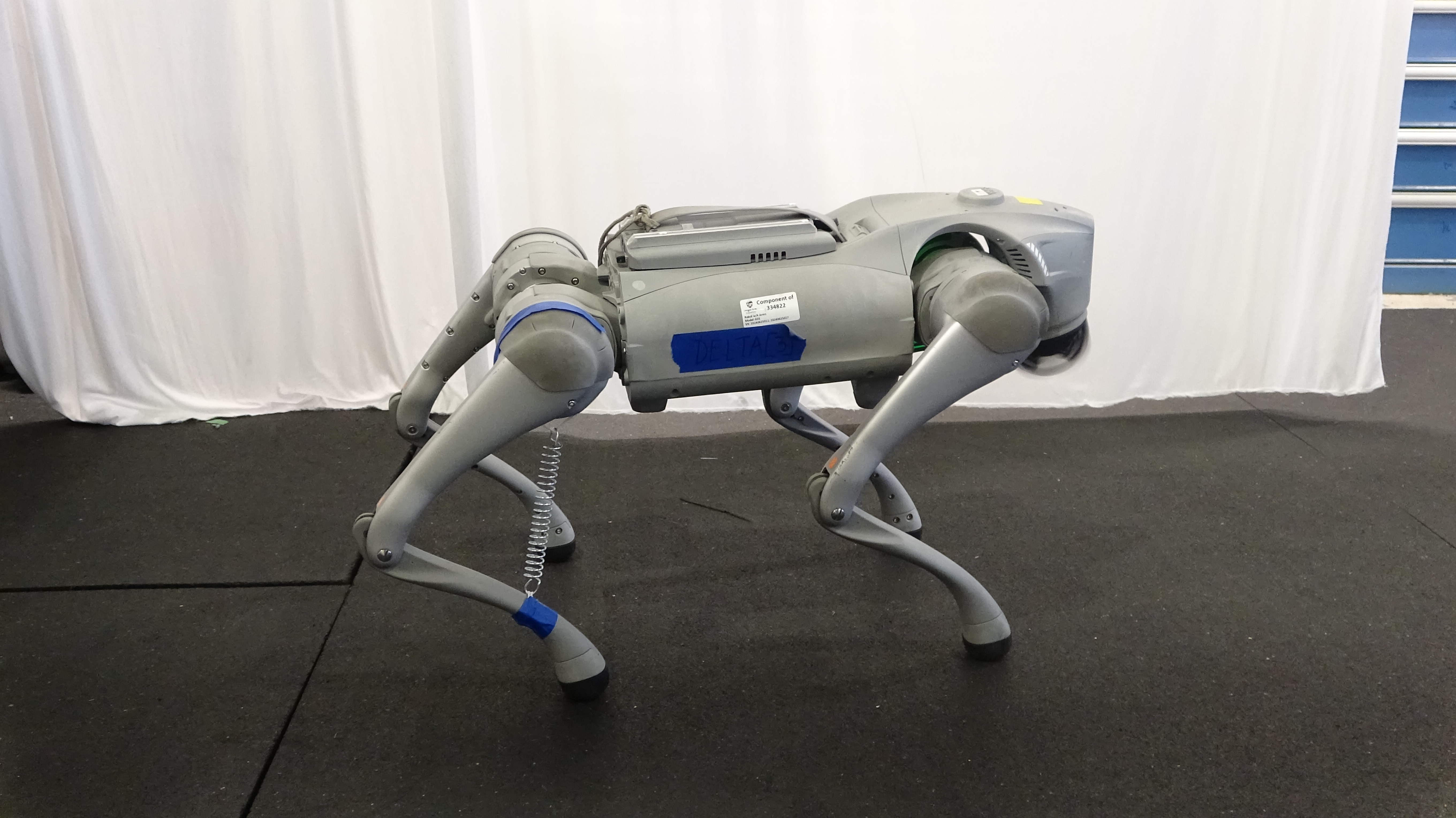}
    \caption{A Unitree Go2 quadruped used in sim-to-real experiments. We tie a spring between the foot and hip joint of the rear right leg, affecting the dynamics of the calf joint. We use this to emulate our \textit{Spring Joint} test on physical hardware.}
    \label{fig:go2_spring_hw}
\end{figure} 
We evaluate our sim-to-real approach on real hardware using two experimental scenarios. The overall procedure mirrors the sim-to-sim pipeline: starting from a simulation-trained locomotion policy $\pi_0$, we collect hardware trajectories, optimize the simulator-modification parameters $\theta$ via \Cref{eq:sim2real_opt} to get $\hat\theta$, and then fine-tune $\pi_0$ with PPO in the modified simulator $\mathcal{M}(\hat\theta)$ to obtain the adapted policy $\pi_1$.

\textbf{Experimental Setup}.
We collect 64 hardware trajectories, each four seconds long, by executing $\pi_0$ on the Go2 robot under random commanded velocities. Since no motion-capture system is available, base position and velocity cannot be measured, and thus none of the state-matching cost functions are applicable. Consequently, all sim-to-real experiments rely exclusively on our proposed Wasserstein metric. Based on our sim-to-sim results, we also adopt the residual actuator model as the simulator-modification function for all sim-to-real experiments.

Because reward and velocity-tracking metrics require privileged signals unavailable on hardware (e.g., contact state, body velocity), we evaluate performance using constant-velocity trials, which support manual performance measurements. For each policy, the robot begins in a static standing pose, receives a constant velocity command for four seconds, and then returns to a zero-command state. For example, for a command of 0.5 m/s, the nominal displacement is 2m along the commanded axis with minimal lateral drift. We evaluate each command dimension (forward velocity, lateral velocity, and yaw rate) independently, using trajectories in both the positive and negative directions. Note that we manually measure the robot’s final $(x,y)$ position, and thus measurement accuracy is approximately $\pm 0.05$m. The accompanying video provides clearer qualitative demonstrations of each policy's behavior on hardware.

\textbf{Spring Joint.}
Our first sim-to-real experiment replicates the \textit{Spring Joint} setting from the sim-to-sim tests. Starting from the same simulation-trained walking policy, we physically attach a spring to the rear-right calf joint (see \Cref{fig:go2_spring_hw}), altering the joint’s effective dynamics. Since only this joint is affected, we train a residual actuator model for that joint alone and then fine-tune the policy in the modified simulator.

The results in \Cref{Table:hw_spring_perf} show substantial improvements in tracking performance. The original policy exhibits over 2m of drift and fails outright for $+y$ sideways walking. In contrast, the fine-tuned policy limits drift to approximately 0.4 in the worst case (roughly 20\% tracking error). 
This demonstrates that the learned residual actuator model provides a significantly more accurate representation of the modified hardware and enables effective finetuning.
Notably, the learned model generalizes beyond the distribution of the collected hardware data. Although the robot could not successfully execute high-speed sideways motion (e.g. +1m/s) during data collection, lower-speed trajectories were sufficient for the sim-to-real pipeline to generalize across the entire command space.
This experiment also demonstrates that our distributional cost function effectively measures the sim-to-real gap even when initial states are not aligned and no controlled resets are possible.
Our method requires only that the same command sequences be used in simulation and hardware, without any additional synchronization.

\textbf{Bipedal Walking}.
Our second experiment evaluates a more challenging scenario, bipedal locomotion as described previously in \Cref{sec:exp_setup}. This represents a real-world use case where there is a noticeable sim-to-real gap betwen the original simulator model and the original unmodified robot. We train a new policy in the nominal simulator $T_{\theta_0}$ to walk on its hind legs using a modified version of the base four-legged walking reward that encourages upright posture, penalizes front-leg contact, and a modified gait reward for two legged walking.

In simulation, the policy exhibits negligible drift across all commanded velocities. However, \Cref{Table:hw_2leg_perf} shows that on real hardware the same policy experiences substantial displacement error, particularly during negative $y$ sideways walking, where significant forward drift occurs.

Training a residual actuator model using the Wass cost and subsequently finetuning the policy again leads to marked improvement. Across command directions, drift decreases substantially, with reductions of up to 50\% in the $-y$ case.
While the hand-measured distances introduce some imprecision due to the robot’s inherently dynamic bipedal stance (the robot cannot stand still), the attached video further illustrates the consistent improvement in tracking behavior.
Overall, the Wasserstein-based residual actuator model yields a simulator that more accurately reflects hardware dynamics, enabling meaningful performance gains on real hardware. These results demonstrate the robustness and practical utility of our proposed sim-to-real pipeline.

\begin{table}[t]
% \vspace{-0.4cm}
\centering
\begin{subtable}{\columnwidth}
\centering
\caption{Original Policy}
\begin{tabular}{l*{3}{c}}
\toprule
 & \makecell{X Axis\\Error (m)} & \makecell{Y Axis\\Error (m)} & \makecell{Yaw Angle\\Error (deg)} \\
\midrule
x +1m/s & -2.0 & 0.483 & 40 \\
x -1m/s & 0.154 & -2.184 & 6 \\
y +1m/s & N/A & N/A & N/A \\
y -1m/s & 0.279 & -0.898 & -6 \\
yaw +0.3925m/s & 0.083 & 0.121 & 3 \\
yaw -0.3925m/s & -0.037 & 0.22 & -11 \\
\bottomrule
\end{tabular}
\end{subtable}

\vspace{0.35em}
\begin{subtable}{\columnwidth}
\centering
\caption{Finetune w/ Residual Actuator Model}
\begin{tabular}{l*{3}{c}}
\toprule
 & \makecell{X Axis\\Error (m)} &\makecell{Y Axis\\Error (m)} & \makecell{Yaw Angle\\Error (deg)} \\
\midrule
x +1m/s & -0.177 & 0.25 & 0 \\
x -1m/s & -0.051 & -.4 & 8 \\
y +1m/s & -0.11 & 0.21 & 22 \\
y -1m/s &  0.2286 & -0.492 & 13 \\
yaw +0.3925m/s & 0.05 & 0.051 & 0 \\
yaw -0.3925m/s & 0.17 & -0.1 & 5 \\
\bottomrule
\end{tabular}
\end{subtable}

\caption{Tracking performance of $\pi_0$ and the finetuned policy $\pi_1$ on hardware for the \textit{Spring Joint} setting. "N/A" means that the policy failed.}
\label{Table:hw_spring_perf}
\vspace{-0.7cm}
\end{table}

\begin{table}[t]
\centering
\begin{subtable}{\columnwidth}
\centering
\caption{Original Policy}
\begin{tabular}{l*{3}{c}}
\toprule
 & \makecell{X Axis\\Error (m)} & \makecell{Y Axis\\Error (m)} & \makecell{Yaw Angle\\Error (deg)} \\
\midrule
x +0.5m/s & 0.6 & 0.35 & 5 \\
x -0.5m/s & 0.0 & -0.4 & 3 \\
y +0.5m/s & 0.2 & -0.25 & 3 \\
y -0.5m/s & 1.2 & -0.5 & 10 \\
yaw +0.3925m/s & 0.06 & 0.17 & 3 \\
yaw -0.3925m/s & 0.0 & 0.025 & -12 \\
\bottomrule
\end{tabular}
\end{subtable}

\vspace{0.35em}
\begin{subtable}{\columnwidth}
\centering
\caption{Finetune w/ Residual Actuator Model}
\begin{tabular}{l*{3}{c}}
\toprule
 & \makecell{X Axis\\Error (m)} & \makecell{Y Axis\\Error (m)} & \makecell{Yaw Angle\\Error (deg)} \\
\midrule
x +0.5m/s & 0.0 & 0.05 & 2 \\
x -0.5m/s & -0.025 & 0.025 & 0 \\
y +0.5m/s & -0.15 & -0.1 & 3 \\
y -0.5m/s & 0.2 & 0.0 & 5 \\
yaw +0.3925m/s & 0.1 & 0.0 & 0 \\
yaw -0.3925m/s & 0.25 & 0.15 & -3 \\
\bottomrule
\end{tabular}
\end{subtable}

\caption{Tracking performance of $\pi_0$ and the finetuned policy $\pi_1$ on hardware for the \textit{Bipedal Walking} setting.}
\label{Table:hw_2leg_perf}
%\vspace{-0.7cm}
\end{table}

\section{CHALLENGES OF HUMANOID APPLICATION}

We also attempted to apply our framework to a more challenging sim-to-real setting on the Agility Robotics Digit humanoid robot. However, we encountered several practical and conceptual difficulties. The most significant discrepancy between simulation and hardware arose in backward walking.
Unlike in the Go2 bipedal walking scenario, where the primary issues were drift or degraded velocity-tracking, our baseline Digit locomotion policy failed outright for backwards walking.
When commanded with negative $x$-velocity, the robot became unstable and consistently fell backward. Consequently, the hardware data we were able to collect contained no valid examples of backward walking, and our sim-to-real pipeline was unable to identify modification parameters that reproduced this backward-falling behavior in simulation. As a result, any policy subsequently finetuned in the adapted simulator reproduced the same failure modes as the original controller.

This exposes an inherent limitation of our approach: it implicitly assumes that the initial sim-to-real transfer is sufficiently successful to generate informative hardware trajectories within the relevant region of the state space. When this assumption does not hold, the hardware data distribution fails to characterize the states in which the mismatch occurs. In such cases, even if our simulator modification procedure identifies modification parameters yielding low matching cost, the underlying sim-to-real gap may persist. Failures present an additional degenerate mode of the optimization problem \Cref{eq:sim2real_opt}. The system can trivially learn modification parameters that cause arbitrary failure in simulation, thereby matching the observed distribution without corresponding to any meaningful or physically plausible change in dynamics. Policies finetuned under such modifications are then systematically misled.

Incorporating failures directly into the dataset does not resolve the issue. Failure states such as falls are highly chaotic and non-deterministic. Once the robot becomes unstable, it may fall in many distinct ways. This variability is exacerbated by the learned control policy itself, which, due to termination conditions during training, is rarely exposed to unrecoverable states and therefore produces extremely noisy and inconsistent actions when failures occur. As a result, trajectories that terminate in superficially similar failure outcomes often exhibit widely differing distributions, making distribution matching ambiguous and unreliable.

At present, our framework lacks a principled mechanism for incorporating such failure regimes into the data distribution or for bridging sim-to-real gaps in regions of state space that are not represented in hardware data. Addressing these limitations will likely require targeted data-collection strategies capable of capturing the causal structure of failures without encountering the degeneracies described above.

\section{CONCLUSIONS}

We presented a distributional sim-to-real metric based on the Wasserstein distance that depends only joint proprioceptive information and requires no time alignment. Combining this with a blackbox optimizer we formed an effective sim-to-real pipeline that can accurately identify model parameters and has comparable matching and finetuned policy performance as stricter time aligned metrics using privileged base information. We then showed the ease and practicallity of our process by fixing sim-to-real gaps without motion capture and less than 5 mintues of real world data. Our pipeline can be generally applied to any robot or simulator modification, allowing for easy general use. We hope that this work will aid others in crossing the sim-to-real gap and enable more complex behaviors to be realized on physical hardware.

Future work may expand upon our process to include learning stochastic dynamics as well. Our current pipeline learns fixed modifications on top of the nominal simulator and then uses regular dynamics randomization when finetuning. One can imagine learning a stochastic/distributional modification (e.g. a distribution of model parameters or having the residual actuator model output a torque distribution to sample from) instead. This would allow for learning a better dynamics randomization as well, instead of just changing a fixed simulator. This would also better fit in real world settings, where real world dynamics can actually be stochastic due to noise, changing motor temperatures, etc.


%%%%%%%%%%%%%%%%%%%%%%%%%%%%%%%%%%%%%%%%%%%%%%%%%%%%%%%%%%%%%%%%%%%%%%%%%%%%%%%%

%%%%%%%%%%%%%%%%%%%%%%%%%%%%%%%%%%%%%%%%%%%%%%%%%%%%%%%%%%%%%%%%%%%%%%%%%%%%%%%%

%%%%%%%%%%%%%%%%%%%%%%%%%%%%%%%%%%%%%%%%%%%%%%%%%%%%%%%%%%%%%%%%%%%%%%%%%%%%%%%%

% \section*{ACKNOWLEDGMENT}

%%%%%%%%%%%%%%%%%%%%%%%%%%%%%%%%%%%%%%%%%%%%%%%%%%%%%%%%%%%%%%%%%%%%%%%%%%%%%%%%

\appendices
\section{Policy Training Details}
We use the same network architecture and optimization hyperparameters for all sim-to-sim and sim-to-real experiments. We represent each of our control policies as a 3 layer MLP with hidden sizes [128, 128, 128] and ELU activations. The policies take in as input joint proprioception position and velocity information, base angular velocity measured from the onboard IMU, a projected gravity vector, a 3 dimensional velocity command vector (x-velocity, y-velocity, yaw-velocity), and the last applied action. They then output 12 joint position targets corresponding to each of the 12 actuated joints. These get passed to an underlying PD controller with fixed gains. The policy is executed at 50Hz, while the PD controller runs at the simulation rate of 200Hz. We use an asymmetrical privileged critic actor-critic setup, where the critic receives true base linear velocity in addition to the regular policy inputs.

We use Proximal Policy Optimization (PPO) to train our policies, with hyperparameters listed in \Cref{table:ppo_params}. Note that the policy's action standard deviation is not fixed, and is an independent learnable parameter. However, during the finetuning stage we keep the standard deviation fixed at 0.5. Policies are trained for 20,000 iterations using 4096 parallel environments, or 1.966 billion samples.

\begin{table}[h]
\centering
\setlength{\tabcolsep}{3pt}
\begin{tabular}{|c|c|}
\hline
Ratio Clipping & 0.2\\
\hline
Entropy Loss Weighting & 0.01\\
\hline
Learning Rate & $1$ x $10^{-3}$\\
\hline
Discount Factor & 0.99\\
\hline
GAE $\lambda$ & 0.95\\
\hline
KL Threshold & 0.01\\
\hline
Maximum Gradient Norm & 1.0\\
\hline
Number of environment steps per policy update & 24\\
\hline
\end{tabular}
\caption{PPO training hyperparameters.}
\label{table:ppo_params}
\end{table}

\section{CMAES Optimization Details}
All CMAES optimizations are run with an initial mean of 0. 

For \textit{FricArm}, the initial covariance is 0.1 for the friction parameters and 0.05 for the armature. All parameters are bound to be between [0, 0.5]. The optimization is allowed to run for 200 iterations.

For the \textit{ActionDelta} and \textit{ResidAct} settings, the initial covariance is set to 0.25 for all network parameters. All parameters are bound between [-1, 1]. The optimization is allowed to run for 10000 iterations. Since these two settings are much higher dimensional, the optimization needs to be run for much longer than in the \textit{FricArm} case for the CMAES $\sigma$ to become sufficiently small and a solution to be settled upon.

% \newpage
% \addtolength{\textheight}{-7cm} %This balances the reference columns. If extra pages are added, reduce this

\def\bibfont{\footnotesize}
\bibliographystyle{IEEEtranN}
\bibliography{bib}

\end{document}